\pdfoutput=1

\documentclass[11pt]{article}

\usepackage[]{ACL2023}

\usepackage{times}
\usepackage{latexsym}

\usepackage[T1]{fontenc}

\usepackage[utf8]{inputenc}

\usepackage{microtype}

\usepackage{inconsolata}

\usepackage{longtable}
\usepackage{color, colortbl}
\definecolor{Gray}{gray}{0.9}
\definecolor{LightCyan}{rgb}{0.88,1,1}
\usepackage{graphicx}
\usepackage{amssymb}
\usepackage{bbding}
\usepackage{booktabs}
\usepackage{float}
\usepackage{subfig}
\usepackage{graphicx}
\usepackage{pgfplots}
\usepackage{afterpage}
\usepackage{makecell}
\usepackage{enumitem}

\usepackage{fontawesome}
\usepackage{scalerel,xparse}

\usetikzlibrary{patterns}
\definecolor{light-gray}{gray}{0.92}

\begin{document}

\title{A Survey of Diffusion Models in Natural Language Processing}


\author{
    Hao Zou, \quad Zae Myung Kim, \quad Dongyeop Kang  \\
    University of Minnesota\\
    \texttt{\{zou00080, kim01756, dongyeop\}@umn.edu}
}

\maketitle
\begin{abstract}
This survey paper provides a comprehensive review of the use of diffusion models in natural language processing (NLP). Diffusion models are a class of mathematical models that aim to capture the diffusion of information or signals across a network or manifold. In NLP, diffusion models have been used in a variety of applications, such as natural language generation, sentiment analysis, topic modeling, and machine translation. This paper discusses the different formulations of diffusion models used in NLP, their strengths and limitations, and their applications. We also perform a thorough comparison between diffusion models and alternative generative models, specifically highlighting the autoregressive (AR) models, while also examining how diverse architectures incorporate the Transformer in conjunction with diffusion models. 
Compared to AR models, diffusion models have significant advantages for parallel generation, text interpolation, token-level controls such as syntactic structures and semantic contents, and robustness. Exploring further permutations of integrating Transformers into diffusion models would be a valuable pursuit. Also, the development of multimodal diffusion models and large-scale diffusion language models with notable capabilities for few-shot learning would be important directions for the future advance of diffusion models in NLP.
\end{abstract}

\section{Introduction}
\begin{figure}[ht]
    \begin{tikzpicture}
    \begin{axis} [ybar stacked,bar width=15pt,height=5cm,width=6.5cm,ylabel={Number of papers},xlabel={Years},symbolic x coords={2021, 2022, 2023}]
    \addplot coordinates {(2021,2) (2022,16) (2023,10)};  
    \addplot[dashed, draw = black, fill = light-gray] coordinates {(2023,40)};
    \end{axis}
    \end{tikzpicture}
\caption{The yearly number of both published and preprinted papers on diffusion models for NLP. For year 2023, the blue bar shows the number collected until the end of April 2023, and the dashed gray bar shows the estimated number for the whole year.}
\label{fig:num_papers}
\end{figure}
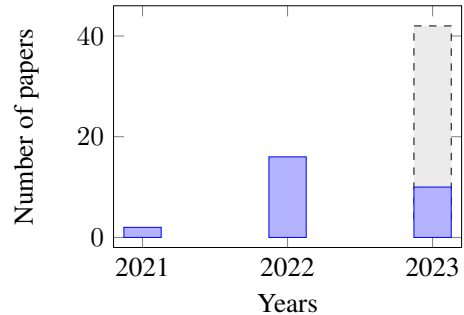

Diffusion models \cite{https://doi.org/10.48550/arxiv.1503.03585, https://doi.org/10.48550/arxiv.2006.11239, https://doi.org/10.48550/arxiv.2010.02502} have shown remarkable performance in image generation and attracted huge attention in the field of artificial intelligence. Researchers have also adopted the models to the field of natural language processing (NLP) and have just started to explore their generative capabilities in the domain (Fig. \ref{fig:num_papers}). To date, diffusion models have been applied to a wide range of generative NLP tasks, such as unconditional text generation, controllable text generation, machine translation, and text simplification.

The main challenge in incorporating diffusion models into NLP is the discreteness of texts, which contrasts with the continuous space in which diffusion is modeled.
To address this challenge, researchers have introduced modifications to the models, and we categorize them into two approaches:
\begin{itemize}[leftmargin=10pt, noitemsep, topsep=0pt]
    \item \textbf{Discrete diffusion models} built on categorical distributions. This method generalizes diffusion process to the discrete domain by corrupting and refining sentences at the token level.
    \item \textbf{Embedding diffusion models} encode discrete texts into continuous space and perform Gaussian noising. As part of this method, additional embedding and rounding steps can be used in the forward and reverse processes, respectively, to convert tokens into embeddings.
\end{itemize}

In the following sections, we first introduce the general framework of vanilla diffusion models and the modified architecture for discrete state spaces in Section \ref{sec:general_framework}. 
In Section \ref{sec:a_categorization_of_diffusion_models_in_nlp}, we classify the surveyed architectures into two aforementioned approaches (discrete vs embedding diffusion models), using specific criteria that have been proposed. 
In Section \ref{sec:diffusion_vs_other_generative_models}, we conduct a detailed comparative analysis of diffusion models against other generative models in NLP domain. Based on empirical evidence, we highlight the advantages of diffusion models over autoregressive (AR) models, specifically in terms of parallel generation, text interpolation, token-level control, and robustness. 
In addition, we explore how various surveyed architectures have incorporated the Transformer with diffusion models for NLP.
We highlight algorithms and techniques proposed for diffusion models in NLP in Section \ref{sec:algorithms_and_techniques}.
Finally, we discuss potential future directions that are both timely and worthy of exploration in Section \ref{sec:challenges_and_future_directions}.

\section{General Framework}
\label{sec:general_framework}

Traditionally, diffusion models have focused on continuous state spaces, but recent advancements have expanded their application to discrete state spaces. Discrete diffusion models operate with discrete variables, such as text or categorical data, which present distinct characteristics and challenges.

A key distinction is the treatment of noise. Continuous diffusion models employ additive Gaussian noise, while discrete diffusion models introduce discrete perturbations or transformations to modify the discrete states. This enables exploration of different states and enhances sample diversity.

Transition probabilities also differ between continuous and discrete diffusion models. Continuous models utilize stochastic differential equations, whereas discrete models define transition probabilities using conditional distributions. These distributions capture dependencies between current and previous states, facilitating information propagation and guiding the diffusion process in discrete state spaces.

\paragraph{Diffusion Models}
Denoising diffusion probabilistic models (DDPMs) were initially introduced by \cite{pmlr-v37-sohl-dickstein15} and enhanced by \cite{https://doi.org/10.48550/arxiv.2006.11239}. DDPMs employ a two-step process: adding Gaussian noise and performing a reverse process to restore the original data. \cite{https://doi.org/10.48550/arxiv.2006.11239} developed DDPMs with an embedding function that maps discrete text to a continuous space, achieving comparable results to state-of-the-art generative models like generative adversarial networks (GANs). Subsequent works \cite{https://doi.org/10.48550/arxiv.2010.02502, https://doi.org/10.48550/arxiv.2105.05233, https://doi.org/10.48550/arxiv.2102.09672, https://doi.org/10.48550/arxiv.2112.10752} have further improved the quality and efficiency of DDPMs.

The forward process generates $X_{t+1}$ by adding noise to $X_t$, creating a dependency solely on $X_t$. This categorizes the diffusion process as a Markov process, where the noise level is determined by the variance $\beta_t \in (0, 1)_{t=1}^T$. The expression for $q(x_t|x_{t-1})$ can be written as follows:
\begin{equation}
    q(x_t|x_{t-1}) = N(x_t; \sqrt{1-\beta_t} \cdot x_{t-1}; \beta_t\mathbf{I})
\end{equation}

By applying the reparameterization approach to depict $X_t$, where $a_t = 1-\beta_t$, $z_t \sim N(0, 1)$, $t \leq 0$, the subsequent result can be obtained:
\begin{equation}
    x_t = \sqrt{\alpha_t} x_{t-1} + \sqrt{1-\alpha_{t}} Z_{t-1}
\end{equation}

When computing $q(x_t|x_0)$, the joint probability distribution of $(x_{1:T}|x_0)$ can be determined because it is established as a Markov chain:
\begin{equation}
    q(x_{1:T}|x_0) = \sum_{t=1}^T q(x_t|x_{t-1})
\end{equation}

Then we can express  $x_t$ at arbitrary time step $t$ with reference to $x_0$ in a closed form, where $\bar \alpha_t = \alpha_1 \alpha_2...\alpha_t$:
\begin{equation}
    q(x_T|x_0) = N(x_t; \sqrt{\bar{\alpha}_t}x_0; (1-\bar{\alpha}_t)\mathbf{I})
\end{equation}

For the reverse process, if we can determine the probability distribution of $x_{t-1}$ based on the given condition of $x_t$, i.e., if $q(x_{t-1}|x_t)$ can be known, then we can iteratively sample random noise to generate an image or sentence. The challenge is to obtain $q(x_{t-1}|x_t)$. To approximate it, we utilize $p_{\theta}(x_{t-1}|x_t)$. Given that the added noise at each step is relatively small, we assume that $p_{\theta}(x_{t-1}|x_t)$ follows a Gaussian distribution that can be modeled using a neural network. The reverse process can be expressed as follows:
\begin{equation}
    p_{\theta}(x_{t-1}|x_t) = N(x_{t-1}; \mu (x_t, t), \sum_\theta (x_t, t))
\end{equation}
\begin{equation}
    p_{\theta}(x_{0:T}) = p(x_T) \sum_{t=1}^T p_{\theta}(x_{t-1}|x_t)
\end{equation}

Applying Bayes' rule, we can express $q(x_{t-1} | x_t, x_0)$ in terms of the known forward conditional probabilities $q(x_t|x_{t-1}, x_0)$, $q(x_{t-1}|x_0)$, and $q(x_t|x_0)$. Our objective is to minimize the mean square error (MSE) loss between the KL divergence of the model $p_{\theta}$ and the true distribution $q$.

\begin{figure*}[h]
\centering
\hspace{-2mm}
\subfloat[Discrete Diffusion Models]{\includegraphics[width=0.5\textwidth]{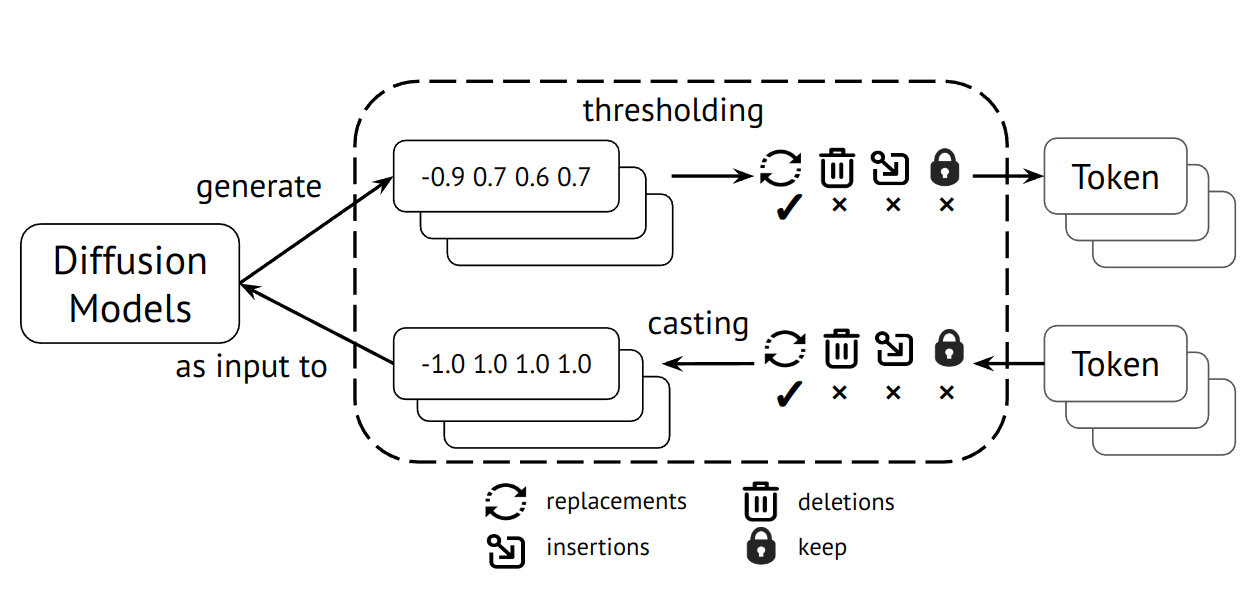}}
\quad
\subfloat[Embedding Diffusion Models]{\includegraphics[width=0.48\textwidth]{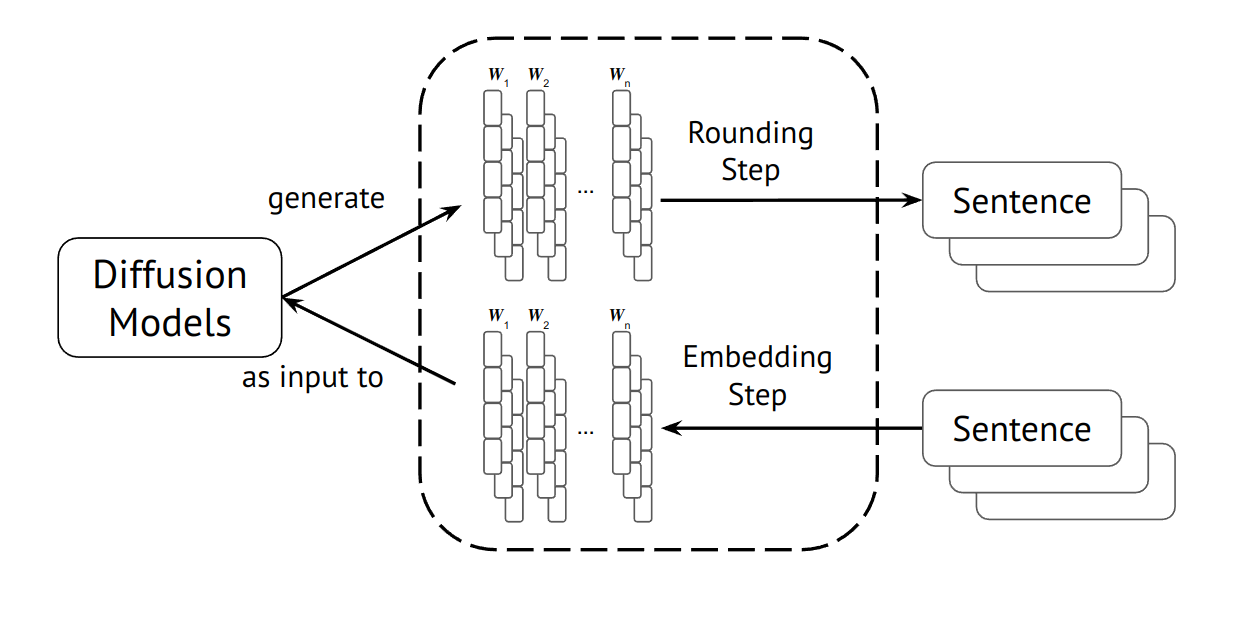}}
\caption{The structures for Discrete Diffusion Models and Embedding Diffusion Models. In Discrete Diffusion Models, the tokens are categorized into categorical values. The figure shows how each token represents a prescribed action to be taken. On the other hand, the Embedding Diffusion Models method involves encoding the entire input sequence into embeddings, followed by applying the diffusion process.}
\label{fig:structures_of_discrete_and_embedding_models}
\end{figure*}

\paragraph{Diffusion models for discrete state spaces}
For scalar discrete random variables with $K$ categories, where $x_t$ and $x_{t-1}$ take values from 1 to $K$, the forward transition probabilities can be represented using matrices. Let $[Q_t]_{i,j} = q(x_t=j|x_{t-1}=i)$. We can denote the one-hot representation of $x$ using a row vector, which can be expressed as follows:
\begin{equation}
   \label{discrete_forward}
    q(x_t|x_{t-1}) = Cat(x_t; p=x_{t-1}Q_t)
\end{equation}
In this context, $Cat(x;p)$ represents a categorical distribution over the one-hot row vector $x$, where the probabilities are determined by the row vector $p$. The term $x_{t-1}Q_t$ corresponds to a row vector-matrix multiplication. An assumption is made that $Q_t$ is independently applied to each pixel of an image or token in a sequence, and that the distribution $q$ factorizes over these higher dimensions as well. Therefore, we can express $q(x_t|x_{t-1})$ in terms of a single element. Starting from $x_0$, we can derive the following $t$-step marginal and posterior at time $t-1$, where $\bar Q_t = Q1Q2...Q_t$:
\begin{equation}
   q(x_t|x_0) = Cat(x_t;p=x_0 \bar Q_t)
\end{equation}
\begin{equation}
   q(x_{t-1}|x_t, x_0) = \frac{q(x_t|x_{t-1}, x_0)q(x_{t-1}|x_0)}{q(x_t|x_0)}
\end{equation}

The Markov property of the forward process ensures that $q(x_t|x_{t-1}, x_0)$ can be simplified to $q(x_t|x_{t-1})$. Similarly, assuming the reverse process $p_\theta(x_t|x_{t-1})$ also exhibits a factorized structure, considering the conditional independence of the image or sequence elements, we can derive the KL divergence between $q$ and $p_\theta$ by aggregating the probabilities across all possible values of each random variable.

\section{A Survey of Diffusion Models in NLP}
\label{sec:a_categorization_of_diffusion_models_in_nlp}

\begin{table*}[hbt!]
\caption{
Comparison of discrete and embedding diffusion models.
} 
\small\centering
\begin{tabular}{@{}p{5cm}|p{3.9cm}|p{2.7cm}|p{3.5cm}@{}}
\toprule
\textbf{Model}  & \textbf{Tasks}  & \textbf{Schedule} & \textbf{Sampling} \\ \midrule
\rowcolor{LightCyan}
\multicolumn{4}{c}{\textbf{Discrete Diffusion Models}} \\ \midrule
Multinomial Diffusion \cite{https://doi.org/10.48550/arxiv.2102.05379} & unconditional text generation, unsupervised spell-checking  & Transition matrices & \textemdash \\ \midrule

D3PM (Discrete Denoising Diffusion Probabilistic Models) \cite{austin2021structured} & char-level text and image generation  & Uniform Transition Matrices & \textemdash \\ \midrule

Zero-shot Diffusion \cite{nachmani2021zeroshot} &  machine translation  & Partial Noising & Classifier-free conditional denoising  \\ \midrule

SUNDAE (Step-unrolled Denoising Autoencoders) \cite{https://doi.org/10.48550/arxiv.2112.06749} & machine translation and unconditional text generation  &  Uniform Transition Matrices & Low-temperature sampling, Argmax-unrolled decoding, fewer token update  \\ \midrule

DiffusionBERT \cite{https://doi.org/10.48550/arxiv.2211.15029} & unconditional text generation  & Spindle & x0-parameterization  \\ \midrule

SSD-LM (Semi-autoregressive Simplex-based Diffusion) \cite{https://doi.org/10.48550/arxiv.2210.17432} & unconditional and controlled text generation   & Logits generation & Greedy projection, Sampling, Multi-hot  \\ \midrule

Bit Diffusion (Generating Discrete Data using Diffusion Models with Self-Conditioning) \cite{chen2023analog} & categorical image generation and image captioning  & \textemdash & Self-Conditioning, Asymmetric Time Intervals \\ \midrule

DiffusER (Discrete Diffusion via Edit-based Reconstruction) \cite{reid2023diffuser} & machine translation, summarization, and style transfer & Edit-based Corruption & Beam Search, 2D Beam Search, Nucleus Sampling  \\ \midrule

Masked-Diffuse LM \cite{chen2023cheaper} & controllable text generation & Mask with Entropy and Reluency & Minimum Bayes Risk \\ \midrule

RDMs (Reparameterized Discrete Diffusion Model) \cite{zheng2023reparameterized} & machine translation & \textemdash & Adaptive Routing Strategy \\ \midrule

\rowcolor{LightCyan}
\multicolumn{4}{c}{\textbf{Embedding Diffusion Models}} \\ \midrule
Diffusion-LM \cite{https://doi.org/10.48550/arxiv.2205.14217} & controllable text generation  & Cosine & Rounding Step and MBR \\ \midrule
	
DiffuSeq \cite{https://doi.org/10.48550/arxiv.2210.08933} & dialogue, question generation, simplification, paraphrasing  &  Partial Noising & Classifier-free Conditional Denoising, MBR  \\ \midrule

SED (Self-conditioned Embedding Diffusion) \cite{strudel2023selfconditioned} & conditional and unconditional text generation, text infilling  & Cosine & Self-conditioning  \\ \midrule

CDCD (Continuous diffusion for categorical data) \cite{https://doi.org/10.48550/arxiv.2211.15089} & prompt completion and infilling, machine translation   & Partial Noising, Time warping & Self-conditioning, Time warping  \\ \midrule

Difformer \cite{https://doi.org/10.48550/arxiv.2212.09412} &  machine translation and abstractive text summarization  & Noise Factor & 2D parallel decoding    \\ \midrule
	
SeqDiffuSeq \cite{https://doi.org/10.48550/arxiv.2212.10325} & dialogue, question generation, simplification, paraphrasing, translation  & Adaptive noise schedule & Self-conditioning  \\ \midrule

DiffuSum \cite{zhang2023diffusum} & extractive text summarization & \textemdash & \textemdash \\ \midrule

GENIE (Diffusion Language Model Pre-training Framework for Text Generation) \cite{lin2023text} & text summarization, common sense generation & \textemdash &  Continuous Paragraph Denoise \\ \midrule

DiNoiSer (Diffused Conditional Sequence Learning by Manipulating Noises) \cite{ye2023dinoiser} & machine translation, text simplification, paraphrasing & Manipulated Noises & Self-conditioning, Condition-enhanced Denoiser, Beam Search, Minimum Bayes Risk\\

\bottomrule

\end{tabular}
\label{tab:categorization}
\end{table*}

We present several studies on diffusion models in NLP by grouping them based on their methods for adapting the diffusion process to the textual domain. Specifically, we have two groups: Discrete Diffusion Models and Embedding Diffusion Models (Figure \ref{fig:structures_of_discrete_and_embedding_models}). The former operates directly in the discrete input space, while the latter involves lifting discrete inputs into a continuous space.

For each category, we then categorize diffusion models into a multi-perspective taxonomy considering the following criteria: (1) the task they are applied to, (ii) schedule methods during the forward process and (iii) sampling methods used for the reverse process. We note that Reluency in ``Schedule'' column indicates a linguistic feature that measures the relevance of word $w$ in one sentence $d$ via tf-idf weights. Entropy is a measurement of the amount of information with entropy $H$ in the word $w$ to reflect the importance of that word.
Table \ref{tab:categorization} shows the categorization.

\subsection{Discrete Diffusion Models}
In the discrete diffusion process, the data is corrupted by switching between discrete values. Discrete diffusion models extend diffusion models to discrete state spaces by corrupting and refining the sentences at the token level.

Multinomial Diffusion \cite{https://doi.org/10.48550/arxiv.2102.05379}  introduces a diffusion-based generative model specifically designed for non-ordinal discrete data. It achieves this by diffusing the data to a uniform categorical distribution, effectively capturing the underlying structure while maintaining controlled randomness. The model's transition mechanism involves independent decisions to either resample or retain values, with resampling performed from a uniform categorical distribution.

D3PMs \citep{austin2021structured} replaces Gaussian noise with Markov transition matrices to diffuse real-world data distribution. It incorporates various types of transition matrices, such as Gaussian kernels, nearest neighbors, and absorbing states, to extend corruption processes. Moreover, D3PMs \citep{austin2021structured} introduces a novel loss function that combines the variational lower bound with an auxiliary cross-entropy loss. Unlike continuous diffusion, D3PMs \citep{austin2021structured} allows precise control over the data corruption and denoising process by selecting $Q_t$ in Equation \ref{discrete_forward}, going beyond the use of additive Gaussian noise.

Zero-Shot Diffusion \cite{nachmani2021zeroshot} utilizes an encoder-decoder architecture with time-based positional encoding for neural machine translation. It employs a transformer encoder to process the source-language sentence and a transformer decoder to handle the noisy target sentence. Notably, this work pioneers conditional text generation using a diffusion model.

Bit Diffusion \cite{chen2023analog} encodes discrete data as binary bits and trains a continuous diffusion model that treats these binary bits as real numbers. It firstly introduces the self-conditioning technique that greatly improves the sample quality and is widely applied to the following works \cite{strudel2023selfconditioned, https://doi.org/10.48550/arxiv.2211.15089, https://doi.org/10.48550/arxiv.2212.10325}. 

SUNDAE \cite{https://doi.org/10.48550/arxiv.2112.06749} proposes step-unrolled text generation and is the first non-AR method to show strong results in both machine translation and unconditional text generation. 

DiffusER \citep{reid2023diffuser} employs a 2-dimensional beam search and edit-based text generation. Instead of a pure end-to-end approach, the system divides the task into edit tagging and generation. It generates a sequence of edits to transform a random noise distribution into high-quality output.

DiffusionBERT \cite{https://doi.org/10.48550/arxiv.2211.15029} combines diffusion models with Pre-trained Language Models (PLMs) \cite{https://doi.org/10.48550/arxiv.1810.04805, https://doi.org/10.48550/arxiv.1910.13461, https://doi.org/10.48550/arxiv.1910.10683, https://doi.org/10.48550/arxiv.2005.14165, Qiu_2020} by training BERT in reverse of a discrete diffusion process. It introduces a new noise schedule for the forward diffusion process and incorporates the time step into BERT \cite{https://doi.org/10.48550/arxiv.1810.04805}. By including the time step, DiffusionBERT captures lost temporal information during diffusion, enhancing the accuracy of the reverse process.

SSD-LM \citep{https://doi.org/10.48550/arxiv.2210.17432} stands out due to two key features. Firstly, it is semi-autoregressive, enabling iterative generation of text blocks and dynamic length adjustment during decoding. Secondly, it is simplex-based, directly applying diffusion on the natural vocabulary space instead of a learned latent space. This approach facilitates the incorporation of classifier guidance and modular control without the need for modifications to existing classifiers.


Masked-Diffuse LM \cite{chen2023cheaper} employs strategic soft-masking, informed by linguistic features, to corrupt both discrete and continuous textual data. It iteratively denoises the data by predicting the categorical distribution. The gradual introduction of perturbations via soft-masking, following an easy-first-generation approach, enhances structural coherence, overall quality, and flexibility in text generation. This pioneering work utilizes linguistic features to effectively corrupt and recover input textual data, improving the generation process.

RDMs \cite{zheng2023reparameterized} introduces a novel reparameterization technique for discrete diffusion models. It employs a stochastic routing mechanism to decide between denoising or noisy resetting for each token. The router ensures uniform processing by assigning equal probabilities to all tokens. This reparameterization simplifies training and enables flexible sampling.

\subsection{Embedding Diffusion Models}
Recent studies \cite{https://doi.org/10.48550/arxiv.2205.14217, https://doi.org/10.48550/arxiv.2210.08933, strudel2023selfconditioned} utilize diffusion processes to generate continuous representations (embeddings) for discrete tokens, known as embedding diffusion models. 

Diffusion-LM \citep{https://doi.org/10.48550/arxiv.2205.14217} constructs diffusion models on continuous word embedding space and incorporates auxiliary losses for joint learning of embedding and network parameters. 

DiffuSeq \citep{https://doi.org/10.48550/arxiv.2210.08933} focuses on sequence-to-sequence generation using encoder-only Transformers and partial noising to define the diffusion process and learn the denoising function.

SED \cite{strudel2023selfconditioned} builds upon the modeling and objectives of Diffusion-LM, introducing a self-conditioning mechanism that enhances baseline performance. Notably, it demonstrates successful scalability to large text datasets like C4 \cite{https://doi.org/10.48550/arxiv.1910.10683}.


Difformer \cite{https://doi.org/10.48550/arxiv.2212.09412} tackles challenges in applying continuous diffusion models to discrete text generation by addressing denoising objective collapse, imbalanced embedding scales, and inadequate noise during training. It introduces three crucial components: an anchor loss function, layer normalization for embeddings, and an increased noise factor to enhance the scale of added noise.

CDCD \cite{https://doi.org/10.48550/arxiv.2211.15089} introduces a variance-exploding stochastic differential equations-based diffusion model tailored for text modeling and machine translation. It integrates time warping, an active learning strategy that dynamically adjusts the noise distribution during training to optimize efficiency.

SeqDiffuSeq \cite{https://doi.org/10.48550/arxiv.2212.10325} incorporates self-conditioning and introduces a method to learn token-level noise schedules for text generation. By leveraging appropriate noise schedules, it aims to enhance the quality of generated samples and likelihood modeling \cite{kingma2021on}. In contrast to DiffuSeq \cite{https://doi.org/10.48550/arxiv.2210.08933}, SeqDiffuSeq \cite{https://doi.org/10.48550/arxiv.2212.10325} explores different model structures and investigates the impact of noise scheduling in sequence-to-sequence tasks.

DiffuSum \cite{zhang2023diffusum} applies diffusion models to enhance extractive summarization. It generates summary sentence representations and extracts relevant sentences using representation matching. The model introduces a contrastive sentence encoding module that employs matching and multi-class contrastive losses to align and diversify representations. Significantly, DiffuSum represents the first known utilization of diffusion models in the field of extractive summarization.

GENIE \cite{lin2023text} is a large-scale diffusion-based language model consisting of an encoder and decoder. It enhances noise removal and paragraph-level coherence through continuous paragraph denoise (CPD) loss in pre-training. The CPD objective guides the diffusion-decoder to reconstruct a clean version of a corrupted text paragraph while preserving semantic and syntactic coherence.

DiNoiSer \cite{ye2023dinoiser} addresses small noise effects on "discrete" embeddings in a continuous space, improving diffusion models through noise manipulation in conditional sequence learning. It tackles the discreteness problem by excluding small-scale noises from diffused sequence learner training. For sampling, it introduces an effective method that consistently indicates large noise scales, enhancing the predictive capabilities by amplifying the influence of source conditions on predictions.

\subsection{Discrete vs. Embedding Diffusion}

\begin{table}[!tp]
\small\centering
\begin{tabular}{@{}cccc@{}}
\toprule
 & \textbf{\makecell{Diffusion \\Process}} & \textbf{\makecell{Classifier-based \\Controls}} & \textbf{\makecell{Refinements \\Adaptation}} \\
\midrule
\makecell{Discrete \\Diffusion} & \makecell{token \\level} & \XSolidBrush & \XSolidBrush \\
\midrule
\makecell{Embedding \\Diffusion} & \makecell{sequence \\level} & \CheckmarkBold &  \CheckmarkBold  \\
\bottomrule
\end{tabular}
\caption{Comparative Analysis of Discrete Diffusion Models and Embedding Diffusion Model. Refinements Adaptation column serves as an indicator of the system's ability to incorporate refinements from continuous diffusion in the image domain.}
\label{tab:discrete_vs_embedding_models}
\vspace{-4mm}
\end{table}

In Table \ref{tab:discrete_vs_embedding_models}, we summarize the advantages of embedding diffusion models over discrete diffusion models. 
\begin{itemize}[leftmargin=10pt, noitemsep, topsep=0pt]

\item \textbf{Diffusion Process}: embedding diffusion models transform discrete inputs into a continuous space, enabling representation of multiple outcomes at intermediate timesteps, particularly crucial in capturing token-level uncertainty in language modeling. In contrast, denoising models operating in the discrete input space lack this ability and are confined to specific tokens.

\item \textbf{Classifier-based Controls}: embedding diffusion models can integrate classifier-based guidance, enhancing the quality of generated samples by leveraging additional information from a classifier to guide the sampling process. In contrast, discrete diffusion models lack this capability, thereby restricting their ability to generate high-quality samples.

\item \textbf{Refinements Adaptation}: \newcite{strudel2023selfconditioned} showed that discrete diffusion approaches do not reap the advantages derived from the advancements made in continuous diffusion methods within the domain of image processing. Conversely, embedding diffusion models exhibit the capacity to leverage these refinements, rendering them more advantageous and valuable in this context.
\end{itemize}

\section{Diffusion vs. Other Generative Models}
\label{sec:diffusion_vs_other_generative_models}
\subsection{Comparison against Latent Variable Models}
Unlike variational autoencoders (VAEs) \cite{kingma2022autoencoding, pmlr-v32-rezende14} or flow-based models \cite{papamakarios2017masked,kingma2018glow}, diffusion models are learned using a fixed procedure with the latent variable having a high dimensionality (same as the original data). GANs \cite{goodfellow2014generative} are known for potentially unstable training and less diverse generations due to their adversarial training nature. VAEs \cite{kingma2022autoencoding, pmlr-v32-rezende14} rely on a surrogate loss. Flow-based models require the construction of specialized architectures to construct reversible transforms.

\begin{table*}[!htp]
\small\centering
\begin{tabular}{@{}ccccccc@{}}
\toprule
& \multicolumn{4}{c}{\textit{Advantages}} & \multicolumn{2}{c}{\textit{Disdvantages}} \\
\midrule
\textbf{Models} & \textbf{\makecell{Parallel \\Generation}} & \textbf{\makecell{Text \\Interpolation}} & \textbf{\makecell{Token-level \\Controls}} & \textbf{\makecell{Robustness to \\Input Corruption}} & \textbf{\makecell{Training\\ Complexity}} & \textbf{\makecell{Model\\Interpretability}}\\
\midrule
AR Models & \XSolidBrush & \XSolidBrush & \XSolidBrush & \XSolidBrush & $\mathcal{O}(n)$ & \faSmileO \\
Diffusion Models & \CheckmarkBold & \CheckmarkBold &  \CheckmarkBold & \CheckmarkBold & $\mathcal{O}(n^T)$ & \faFrownO \\

\bottomrule
\end{tabular}
\caption{Comparative Analysis of Autoregressive (AR) and Diffusion Models in NLP. Token-level Controls of diffusion models include syntactic structure, parse trees, semantic content, parts-of-speech, etc. In terms of training complexity, diffusion models employ multiple rounds of diffusion steps $T$ to generate the entire sequence. Each diffusion step involves optimizing the objective function to capture the denoising process. Specifically, Transformer models are utilized to model the denoising process within each diffusion step.}
\label{tab:AR_vs_diffusion_models}
\end{table*}

As \newcite{https://doi.org/10.48550/arxiv.2211.15089} notes, diffusion models have a distinct advantage over models like VAEs and GANs, which generate data in a single forward pass. Diffusion models instead focus on reconstructing a small amount of information that has been removed by the corruption process, making the task less challenging.

\subsection{Comparison against Autoregressive Models}

\begin{figure}[t]
\centering
\subfloat[Diffusion-based language model ]{\includegraphics[width=1.0\linewidth]{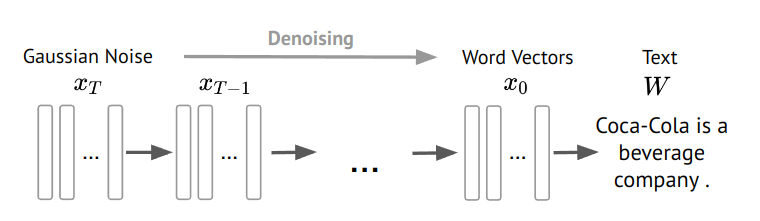}}

\subfloat[AR language model]{\includegraphics[width=0.8\linewidth]{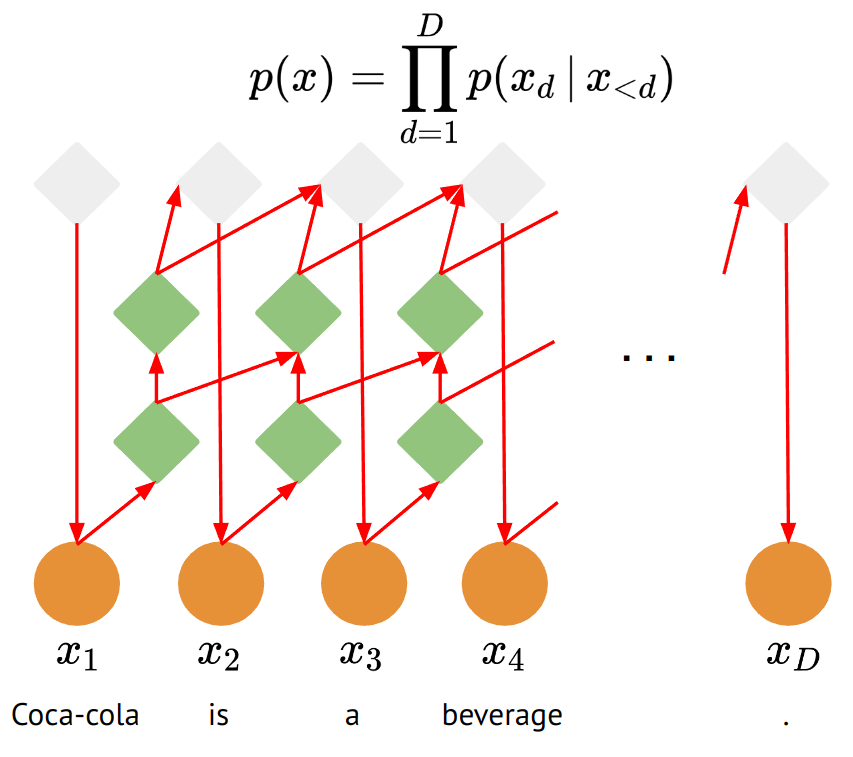}}
\caption{The comparison between diffusion-based language models and autoregressive language models: diffusion LM iteratively denoises a sequence of Gaussian vectors into word vectors, while AR language model predicts the next word in a sequence of words based on the previous predictions.}
\label{fig:AR_vs_diffusionlm}
\end{figure}

\begin{figure*}[!htbp]
\centering\small
\includegraphics[width=0.9\textwidth]{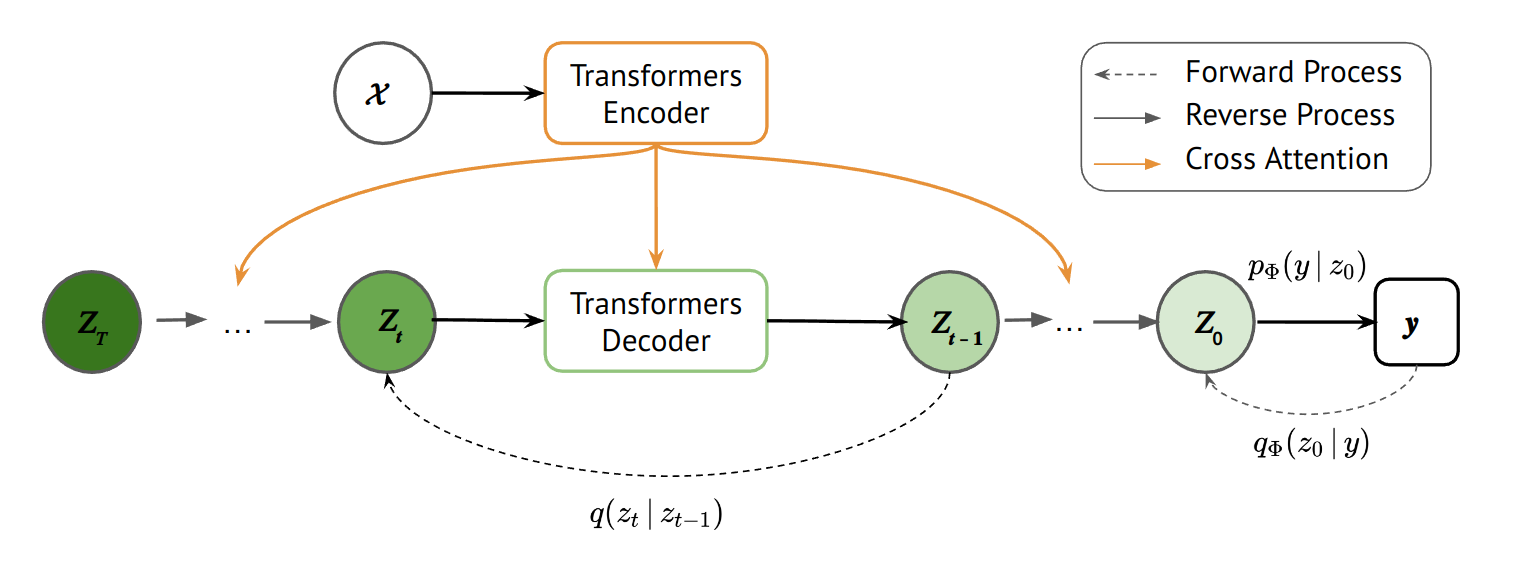}
\caption{Illustration for how to incorporate Transformers architecture with diffusion models in NLP.}
\label{fig:transformer_in_diffusion_models}
\vspace{-4mm}
\end{figure*}
\begin{table*}[h!]
\small\centering
\begin{tabular}{@{}ccccc@{}}
\toprule
\textbf{Architecture} & \textbf{Training Corpus} & \textbf{Parameter Size} & \textbf{with Transformer} & \textbf{Pre-trained} \\
\midrule
\rowcolor{LightCyan}
\multicolumn{5}{c}{\textbf{Discrete Diffusion Models}}\\
\midrule
Multinomial Diffusion & text8, enwik8 & \textemdash & 12-layer Transformer & \XSolidBrush\\
\midrule
D3PMs & text8, LM1B, CIFAR-10 & \textemdash & 12-layer Transformer & \XSolidBrush\\
\midrule
Zero-shot Diffusion & \makecell{WMT14 (DE-EN, FR-EN) \\ WMT19 (DE-FR)} & \textemdash & 12-layer Transformer & \XSolidBrush\\
\midrule
SUNDAE & \makecell{WMT14 (EN-DE)\\C4 \\ Python code dataset} & 63M & \makecell{encoder-decoder Transformer\\ causality masking removed \\in the decoder} & \XSolidBrush\\
\midrule
DiffusionBERT & LM1B & 110M & \makecell{bert-base uncased is \\trained for reverse process} & \CheckmarkBold\\
\midrule
SSD-LM & OpenWebText & 0.4B & \makecell{bi-directional \\Transformer encoder\\a timestep embedding added\\
before the first \\Transformer block} & \XSolidBrush\\
\midrule
Bit Diffusion & \makecell{CIFAR-10, ImageNET\\ MSCOCO 2017} & \textemdash & 6-layer Transformer decoder & \XSolidBrush\\
\midrule
DiffusER & \makecell{WMT’14\\ CNN/DailyMail, Yelp} & \textemdash & \makecell{6-layer Transformer to predict\\ the edit operations\\6-layer Transformer for generator} &  \XSolidBrush\\
\midrule
Masked-Diffuse LM & E2E & 80M & \makecell{BERT to encode the input text\\Transformer to module \\the reverse process} & \CheckmarkBold \\
\midrule
RDMs \cite{zheng2023reparameterized} & \makecell{IWSLT14 (DE-EN)\\WMT14 (EN-DE)\\WMT16 (EN-RO)} & \textemdash & \makecell{Length prediction module\\on top of Transformer encoder} & \CheckmarkBold\\
\midrule
\rowcolor{LightCyan}
\multicolumn{5}{c}{\textbf{Embedding Diffusion Models}}\\
\midrule
Diffusion-LM & E2E, ROCStories & 80M & 12-layer Transformer & \XSolidBrush \\ 
\midrule
DiffuSeq &  \makecell{CCD,  Quasar-T\\Newsela-Auto \\ Wiki-Auto, QQP} & 91M & 12-layer Transformer & \XSolidBrush\\ 
\midrule
SED & C4 & 135M \& 420M & 12-layer Transformer & \CheckmarkBold\\ 
\midrule
CDCD & \makecell{MassiveText, C4\\WMT2014, WMT2020} & 1.3B & Mask-conditional Transformer & \CheckmarkBold \\ 
\midrule
Difformer &  \makecell{IWSLT14, WMT14\\ WMT16, Gigaword} & \textemdash  & 6-layer Transformer & \XSolidBrush\\ 
\midrule
SeqDiffuSeq & \makecell{CCD, Quasar-T, Wiki-Auto \\ QQP, IWSLT14} & \textemdash & 12-layer Transformer & \XSolidBrush \\ 
\midrule
DiffuSum & \makecell{CNN/DailyMail,
XSum\\ PubMed} & 13M & \makecell{8-layer Transformer as encoder\\12-layer Transformer as generator} & \CheckmarkBold\\
\midrule
GENIE & \makecell{Gigaword, CNN/DailyMail\\XSum, CommonGen}  & \textemdash & \makecell{6-layer Transformer as encoder\\6-layer cross attention Transformer\\ as
denoising architecture} & \CheckmarkBold\\
\midrule
DiNoiSer &  \makecell{IWSLT14 (DE-EN)\\ WMT14 (EN-DE, EN-RO)\\Wiki-Auto, QQP} & \textemdash & 12-layer Transformer & \XSolidBrush\\

\bottomrule
\end{tabular}
\caption{Training Corpus and connection with Transformer for Discrete and Embedding Diffusion Models. Parameter column refers to the size of used Transformer architecture specifically. Pre-trained column indicates whether the system uses the pre-trained word embedding or not.}
\label{tab:connection_with_transformer}
\end{table*}

Autoregressive (AR) models currently dominate the field of language modeling. Also known as causal modeling or the next-token prediction task, AR modeling learns the joint distribution over a token sequence $p(x_1, x_2, ..., x_N)$ by factorizing it into sequential conditionals $p(x_k|x_1, ..., x_{k-1})$ and model them separately with shared parameters (see Figure \ref{fig:AR_vs_diffusionlm}). This means that sampling always proceeds along the left-to-right direction of the sequence. However, in many cases, the ability to go back and refine the earlier parts of the sequence should be useful. In Figure \ref{fig:AR_vs_diffusionlm}, we illustrate the fundamental distinctions between AR and diffusion models, and highlight the distinctive features of the diffusion architecture that endow it with the ability to refine the previous generations, which has potentials to advance the state-of-the-art in the field. 

Additionally, \newcite{strudel2023selfconditioned} reveals that, compared to AR models \cite{Bengio2003ANP, Sutskever2011GeneratingTW, Austin2021StructuredDD, Hoffmann2022TrainingCL}, diffusion models can predict all tokens in a sequence at once, which increases interactions between tokens, potentially leading to more coherent samples. Similarly, \newcite{https://doi.org/10.48550/arxiv.2112.06749} and \newcite{https://doi.org/10.48550/arxiv.2205.14217} note that the fixed generation order (left-to-right) from AR models limits the model's flexibility in many controllable generation settings. For example, infilling task, which imposes lexical control on the right contexts, and the syntactic structure control task, which controls global properties involving both left and right contexts. More importantly, this prohibits the iterative refinement of complete text drafts from making them more self-consistent, which is a common task for human writers.

In Table \ref{tab:AR_vs_diffusion_models}, we summarize the empirical benefits of diffusion models over AR models. We categorize them into four aspects: parallel generation, sentence interpolation, token-level control, and robustness to input corruption. 
\begin{itemize}[leftmargin=10pt, noitemsep, topsep=0pt]
    \item \textbf{Parallel Generation}: diffusion models exhibit a notable departure from the autoregressive nature of AR models. While AR models generate output tokens sequentially conditioned on preceding tokens, diffusion models adopt a parallel generation approach, enabling simultaneous generation of all output tokens. This characteristic enhances the speed and efficiency of text generation, rendering diffusion models particularly suitable for real-time applications. 
    \item \textbf{Text Interpolation}: diffusion models demonstrate a superior capacity for text interpolation. Leveraging the denoising process inherent in their design, diffusion models can generate intermediate sentences between two given sentences, ensuring smooth transitions and coherent outputs. This capability enhances the overall fluency and cohesiveness of generated text.
    \item \textbf{Token-level Controls}: Diffusion models provide advanced Token-level Controls, facilitating fine-grained manipulation of generated outputs. This level of control enables precise modifications and interventions in the generated sequences, enhancing the interpretability and applicability of diffusion models in diverse downstream tasks.
    \item \textbf{Robustness to Input Corruption}: Diffusion models exhibit enhanced robustness due to their denoising mechanism that facilitates the reconstruction of the original input. This process aids in mitigating errors and noise present in the input sequence. Consequently, diffusion models are capable of capturing a broader spectrum of input variations by learning a more adaptable distribution over the input data.
\end{itemize}


In summary, diffusion models offer empirical advantages over AR models, encompassing parallel generation, text interpolation, and advanced token-level controls. These characteristics underscore the potential of diffusion models in various text generation scenarios, emphasizing their efficiency, coherency, and flexibility.
In addition to the advantages discussed in Table \ref{tab:AR_vs_diffusion_models}, we also identify two significant disadvantages of diffusion models compared to AR models in terms of training complexity and interpretability.
\begin{itemize}[leftmargin=10pt, noitemsep, topsep=0pt]
    \item \textbf{Training Complexity}: Diffusion models are more difficult to train than AR models due to their more complex architecture and optimization objective. In a diffusion model, the entire sequence is generated simultaneously through multiple rounds of diffusion steps, which involve applying a non-linear function to a set of latent variables to obtain the next generation of the sequence. This requires optimizing a complex objective function that includes both the data likelihood and the distance between the generated and ground-truth sequences. On the other hand, AR models generate sequences sequentially by conditioning each time step on the previous ones. This allows for a simpler optimization objective and faster convergence during training.
    \item \textbf{Model Interpretability}: Diffusion models involve multiple non-linear transformations during the diffusion process, resulting in abstract representations in the latent space. These representations may not have a clear interpretation or meaning, and understanding how a specific output sequence is generated from the input can be challenging. This makes diffusion models less interpretable. In contrast, AR models generate sequences step by step, building on the previous steps. Each step is influenced by the preceding steps, making it easier to understand how the output sequence is generated based on the input. AR models are more interpretable due to this sequential nature.
These observations highlight the trade-offs associated with diffusion models, emphasizing the need to consider both their advantages and disadvantages in practical applications.
\end{itemize}

\subsection{Transormers with diffusion models}
Transformers architecture could be combined with diffusion models, as depicted in Figure \ref{fig:transformer_in_diffusion_models}. Specifically, the Transformer models are used in the encoder-decoder layout to model the denoising function. During the reverse process, the input sequence $x$ therefore only requires one forward computation.

Furthermore, Table \ref{tab:connection_with_transformer} provides a comprehensive summary of the training corpus of surveyed systems, highlighting their associations with Transformers. This includes details such as the parameter size and the specific architectures employed by each system for modeling denoising functions, as well as their utilization of pre-trained representations from Transformers during the diffusion process. We hope that this summary can provide researchers with rapid insights into the interplay between Transformers and diffusion models in NLP.

\begin{figure*}[!ht]
\centering\small
{\includegraphics[width=1.0\textwidth]{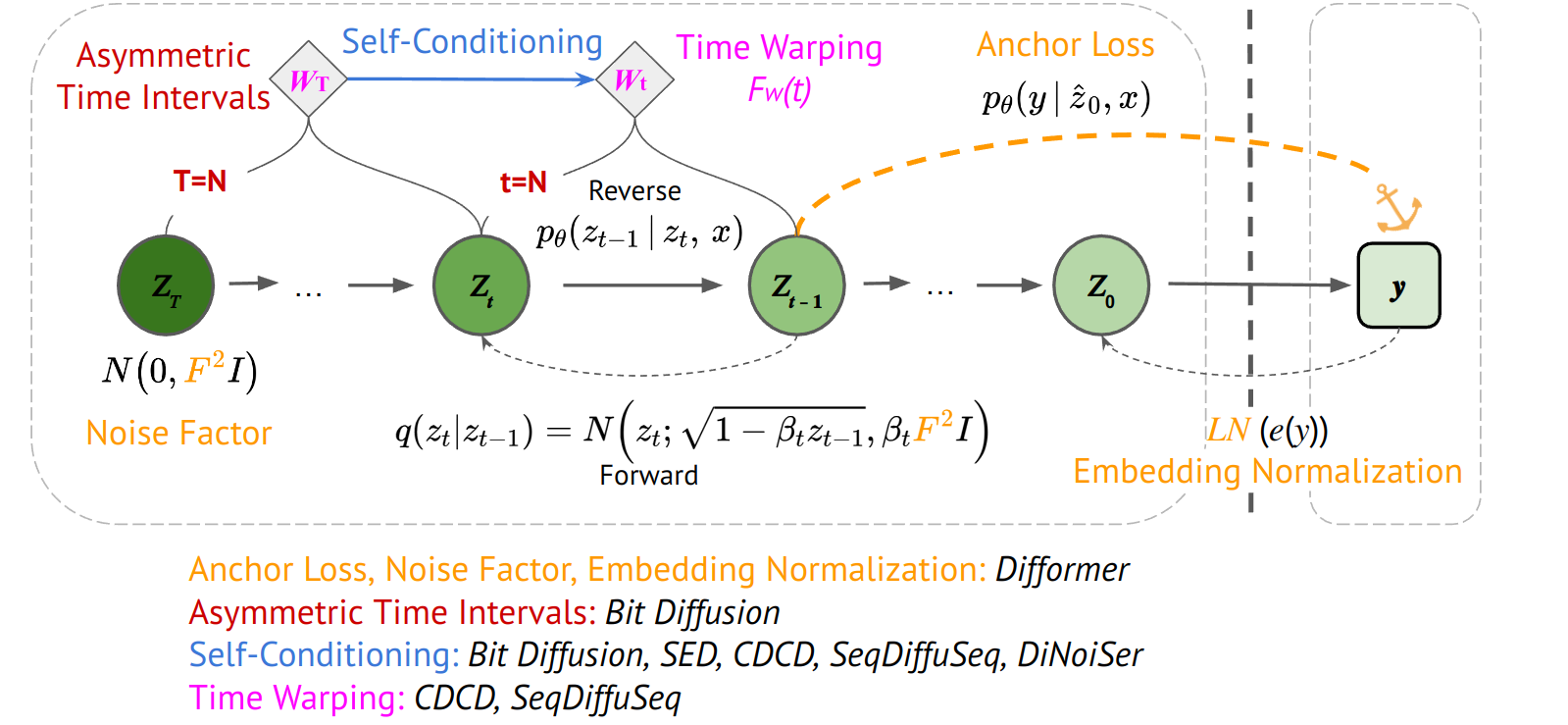}}
\caption{Algorithms proposed to adapt the discrete data. Details of the proposed architectures are described in Section \ref{sec:a_categorization_of_diffusion_models_in_nlp}. Details of the algorithms are described in Section \ref{sec:algorithms_and_techniques}. }
\label{fig:different_types_of_diffusion_models}
\end{figure*}

\section{Algorithms \& Techniques}
\label{sec:algorithms_and_techniques}

In this section, we highlight algorithms and techniques proposed for diffusion models in NLP. They are twofold: (1) adapting the models to discrete variables and (2) improving sampling procedures. Figure \ref{fig:different_types_of_diffusion_models} depicts the algorithms proposed from the surveyed papers.

\subsection{Adapting Discrete Variables}
\subsubsection{Diffusion Steps}
To optimize the objective function, DDPM \citep{https://doi.org/10.48550/arxiv.2006.11239} utilizes the property that the noise added at each time step in the diffusion process is Gaussian noise; hence the concrete expressions of the objective can be derived. However, the Gaussian distribution here is mainly for continuous domains such as image generations. Hence, D3PM \citep{https://doi.org/10.48550/arxiv.2107.03006} proposed a new method for adding noises for discrete variables. D3PM defined a series of transition matrices that transformed the discrete tokens into \verb|[MASK]| based on pre-defined probabilities at different time steps. 

\subsubsection{Objective Functions}
\paragraph{Predicting initial inputs directly}
Traditionally, for the approximations of the mean values of each time step, DDPM \citep{https://doi.org/10.48550/arxiv.2006.11239} predicts the noise at each time step directly, however, Diffusion-LM \citep{https://doi.org/10.48550/arxiv.2205.14217} found that the model might fail to generate the initial input $x_0$ that commits to a single word as the denoising steps cannot ensure that $x_0$ lies precisely on the embedding of a word. To solve this problem, Diffusion-LM \citep{https://doi.org/10.48550/arxiv.2205.14217} predicts the initial input $x_0$ directly in their objective functions. 

\paragraph{Partial noising and conditional denoising}
DiffuSeq \citep{https://doi.org/10.48550/arxiv.2210.08933} connects the conditional text $c$ and the target text $x$, and adds noise only to the target text $x$ in forward process while denoising only $x$ in the denoising process. In contrast to Diffusion-LM's approach \citep{https://doi.org/10.48550/arxiv.2205.14217}  of classifier-guided diffusion, DiffuSeq \citep{https://doi.org/10.48550/arxiv.2210.08933} employs a method of classifier-free diffusion that is directed by spatial points. Thus, the system is capable of producing conditional generations in the absence of external classifiers.

\subsection{Sampling from Latent Space}
\paragraph{Asymmetric Time Intervals}
Time step plays a critical role in diffusion models. During typical reverse diffusion, symmetric time intervals are often used for both state transition and time reduction, resulting in shared $t$ for $f(x_t, t)$. However, \newcite{chen2023analog} shows experimentally that when taking a larger step, using asymmetric time intervals with $f(x_t, t')$, implemented via a simple manipulation of time scheduling at generation, can lead to improved sample quality.

\paragraph{Self-Conditioning}
When estimating the data sample by the denoising network $f$ at a time step, conditioning the network directly on its previously estimated samples (as opposed to discarding them) can provide better sample quality \citep{chen2023analog}.

\paragraph{Time Warping}
\newcite{https://doi.org/10.48550/arxiv.2211.15089} introduces time warping, an active learning strategy that automatically adapts the distribution of noise levels sampled during training to maximize efficiency. The method alters the relative weighting of the noise levels corresponding to different time steps $t$. 
To sample $t$ non-uniformly in practice, the inverse transform sampling can be used: first generate uniform samples $u$ $\in$ $[0, 1]$ and then warp them using the inverse cumulative distribution function (CDF) of the distribution which corresponds to the desired weighting: $t$ = $F - (u)$. This time warping procedure is equivalent to time reweighting in expectation, but more statistically efficient.

\section{Challenges \& Future Directions}
\label{sec:challenges_and_future_directions}
\begin{figure}[!t]
\centering\small
{\includegraphics[width=1.0\linewidth]{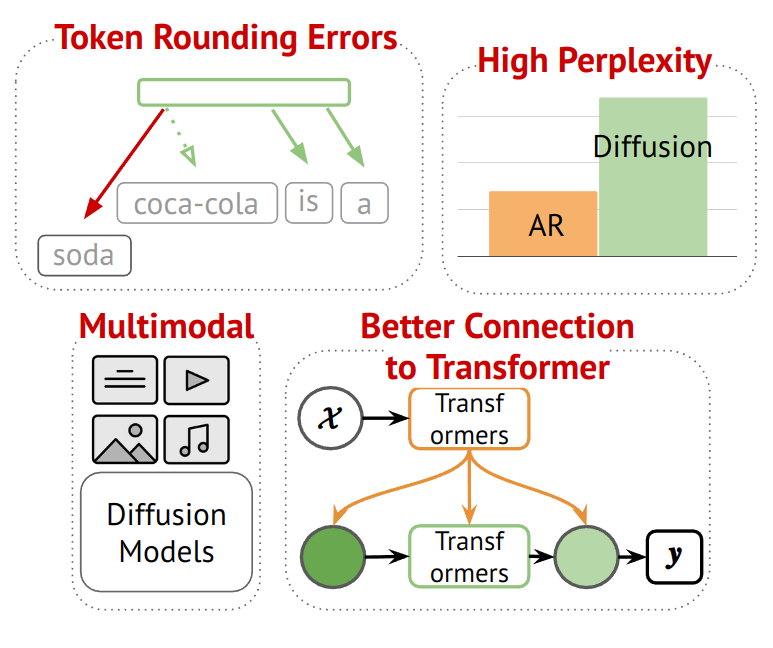}}
\caption{Challenges and future directions we conclude based on the surveyed papers.}
\label{fig:challenges_and_future_directions}
\end{figure}

In this section, we advance potential lines of inquiry that are both contemporarily significant and intellectually deserving of investigation (Figure \ref{fig:challenges_and_future_directions}).

\subsection{General Challenges}
\paragraph{Latent Space Restriction} Diffusion models impose a restriction on the latent space representations, as the dimensions of latent vectors and inputs must be the same. This constraint limits the representational power of the latent vector.

\paragraph{Computational Cost} The convergence of diffusion models requires a large number of iterations, which can lead to significant computational costs, especially when dealing with large datasets.

\paragraph{Sensitivity} Diffusion models can be very sensitive to the choice of hyperparameters, such as diffusion coefficient, time step size, number of diffusion steps, etc., which can lead to suboptimal performance or even failure to converge.

\paragraph{Dependence on diffusion process assumptions} Diffusion models rely on the assumption that information diffuses smoothly and uniformly across the data, which may not always hold in practice. Given perfect mathematical formulation, the diffusion process itself might not be intuitive enough. For instance, optimizing from a totally noisy distribution is quite different to human mind.

\paragraph{Limited interpretability and explainabilities} The black-box nature of diffusion models makes it challenging to understand how they make decisions, limiting their interpretability. For instance, the latent vectors learned from diffusion models do not have any linguistic or structural explainabilities.

\subsection{NLP-Specific Challenges}
\paragraph{Token Rounding Errors}
The learned embeddings through embedding diffusion models define a mapping from discrete text to the continuous $x_0$. We now describe the inverse process of rounding a predicted $x_0$ back to discrete text. Rounding is achieved by choosing the most probable word for each position. However, empirically, the model
fails to generate $x_0$ that commits to a single word \cite{https://doi.org/10.48550/arxiv.2205.14217}.

\paragraph{High Perplexity}
As stated in \newcite{https://doi.org/10.48550/arxiv.2205.14217, lovelace2022latent}, the perplexity from diffusion models lags behind AR models. However, measuring perplexity with a pretrained AR model such as GPT-2 may bias the metric towards AR models. Besides, previous studies have demonstrated that generating text with low perplexity does not necessarily imply high quality, but rather suggests degenerate behavior. \cite{nadeem-etal-2020-systematic, zhang-etal-2021-trading}. Hence, better metrics which have a stronger correlation with human judgements of quality are needed. For this factor, \newcite{pillutla2021mauve} proposed MAUVE Score,  a metric for open-ended text generation that compares the distribution of generated text with that of reference text using divergence frontiers, to better correlate with human judgments.

\subsection{Potential Future Directions}
\paragraph{More Advanced Ways to connect Transformers}  How to better combine the spatiality of Transformer and temporality of Diffusion is a tricky question since the ideologies for Transformer and Diffusion are from totally different perspectives. Common architectures from our surveyed paper make The time step $t$ included in the neural net through a Transformer sinusoidal position embedding in each block. And currently people just diffuse the whole sequence of the sentences, diffusion process on single token might be interesting to try on. More variations of injecting Transformers into Diffusion might be needed to explore and deeper analysis is needed with strong foundations. 

\paragraph{Large Scaled Diffusion Language Models with impressive few-shot learning capabilities}
Giant language modeling has made significant strides in recent years and has become a dominant area of research in artificial intelligence. With advances in deep learning and natural language processing, large language models like GPT-3 have shown impressive abilities in tasks such as language translation, text generation, question-answering, and even programming. Currently only SED \cite{strudel2023selfconditioned} has studied the scaling issues for diffusion models in NLP, the enormous potential of Large-Scale Diffusion Language Modeling in few-shot learning warrants further exploration.

\paragraph{Multimodal Diffusion Modeling}
In recent years, there has been a growing interest in developing visual language models (VLMs), which are deep learning models that can understand the relationship between images and natural language. The amazing few-shot performance of VLMs shows great potential to transform how machines interact with the visual world and language, such as Vision-Language Pre-training (ViLBERT) model from Facebook AI Research (FAIR) and the Georgia Institute of Technology, and Flamingo from DeepMind. However, current VLMs are all based on Transformers, the incorporation of Diffusion Models presents vast potential for exploration and discovery.

\section{Conclusion}

This survey paper extensively discusses the formulations, strengths, limitations, and applications of diffusion models in NLP. We conduct a comprehensive comparison between diffusion models and alternative generative models, focusing on autoregressive (AR) models. Additionally, we explore the integration of the Transformer architecture with diffusion models across various architectures.

Our findings demonstrate the significant advantages of diffusion models over AR models. They excel in parallel generation, enabling faster and more efficient text generation. Diffusion models also demonstrate superior performance in sentence interpolation, token-level controls, and robustness to input corruption. Further research on integrating Transformers into diffusion models and developing multimodal and large-scale diffusion language models for few-shot learning is crucial.

In summary, this survey paper provides a comprehensive overview of diffusion models in NLP, highlighting their benefits, comparative analysis with AR models, and avenues for future research. We hope it can contribute to the understanding and advancement of diffusion models in the field of NLP.

\section*{Limitations}
The selection of diffusion models included in this paper may introduce a bias based on our knowledge and availability of resources. This could potentially exclude relevant diffusion models that were not considered or well-known at the time of the survey. It is crucial to acknowledge that the selection of specific models and the exclusion of others can impact the comprehensiveness and generalizability of the findings. Another limitation pertains to the understanding and interpretation of the inner workings and decision-making processes of the surveyed diffusion models. Diffusion models in NLP, particularly those employing deep learning techniques, are often regarded as black-box models with limited interpretability. The lack of interpretability can impede the trust and acceptance of diffusion models in practical applications.

\section*{Ethics Statement}
Diffusion models in NLP may be influenced by biases present in the training data, highlighting the need to consider the ethical implications of deploying biased models in real-world applications. Furthermore, the impact of diffusion models in NLP extends to shaping public opinion, influencing decision-making processes, and affecting social dynamics. Therefore, we prioritize responsible use and communication of the findings in this paper, avoiding sensationalism, misrepresentation, or overgeneralization of the capabilities and limitations of diffusion models in NLP to ensure a well-rounded understanding among the public.

\section*{Acknowledgement}
This project is supported in part by Sony Research Grant.

\bibliography{anthology,custom}
\bibliographystyle{acl_natbib}

\end{document}


\maketitle


\clearpage
\onecolumn
\small\centering
\begin{longtable}{|p{2cm}|p{3cm}|p{2.8cm}|p{2cm}|p{3.5cm}|}
\caption{To classify existing models, we consider three criteria: the
task, the denoising condition, and the underlying approach (architecture). Additionally, we list the data sets and evaluation metrics on which the surveyed models are applied. We use the following abbreviations in the architecture column: D3PM (Discrete Denoising Diffusion Probabilistic Models), SUNDAE (Step-unrolled Denoising Autoencoders for Text Generation), DiffusionBERT (Improving Generative Masked Language Models with Diffusion Models), SSD-LM (Semi-autoregressive Simplex-based Diffusion Language Model for Text Generation and Modular Control), Bit Diffusion (Generating Discrete Data using Diffusion Models with Self-Conditioning), DiffusER (Discrete Diffusion via Edit-based Reconstruction), SED (Self-conditioned Embedding Diffusion for Text Generation), CDCD (Continuous diffusion for categorical data)} \label{tab:maintable} \\

\hline \multicolumn{1}{|c|}{\textbf{paper}} & \multicolumn{1}{c|}{\textbf{Tasks}} & \multicolumn{1}{c|}{\textbf{Denoising Condition}} & \multicolumn{1}{c|}{\textbf{Architecture}} & \multicolumn{1}{c|}{\textbf{Datasets}}  \\ \hline 
\endfirsthead

\multicolumn{ 5}{|c|}{\textbf{Discrete Diffusion Modes}} \\ \hline

\rowcolor{Gray} 
\newcite{austin2021structured} & char-level text and image generation & unconditional & D3PMs & text8, LM1B, CIFAR-10 \\ \hline

\rowcolor{Gray} 
\newcite{https://doi.org/10.48550/arxiv.2102.05379} & unconditional text generation and unsupervised spell-checking & unconditional & Multinomial Diffusion & text8, enwik8 \\ \hline

\rowcolor{Gray} 
\newcite{https://doi.org/10.48550/arxiv.2112.06749} & machine translation and unconditional text generation & unconditional & SUNDAE & WMT’14, C4,  a dataset of Python code \\ \hline

\rowcolor{Gray} 
\newcite{https://doi.org/10.48550/arxiv.2211.15029} & unconditional text generation  & unconditional & DiffusionBERT & LM1B \\ \hline

\rowcolor{Gray} 
\newcite{https://doi.org/10.48550/arxiv.2210.17432} & unconditional text generation and sentiment controlled text generation & unconditional & SSD-LM & OpenWebText \\ \hline

\rowcolor{Gray} 
\newcite{chen2023analog} & categorical image generations and image captioning & unconditional & Bit Diffusion & CIFAR-10, ImageNET; MSCOCO 2017 \\ \hline

\rowcolor{Gray} 
\newcite{reid2023diffuser} & machine translation, summarization, and style transfer & unconditional & DiffusER & WMT’14, CNN/DailyMail, Yelp  \\ \hline

\multicolumn{ 5}{|c|}{\textbf{Embedding Diffusion Modes}} \\ \hline
	
\rowcolor{LightCyan}
\newcite{https://doi.org/10.48550/arxiv.2205.14217} & controllable text generation & unconditional, classifier guidance & Diffusion-LM & E2E, ROCStories \\ \hline

\rowcolor{LightCyan}
\newcite{https://doi.org/10.48550/arxiv.2210.08933} & sequence to sequence text generation & classifier-free conditional & DiffuSeq &  CCD,  Quasar-T, Newsela-Auto and Wiki-Auto, QQP \\ \hline

\rowcolor{LightCyan}
\newcite{https://doi.org/10.48550/arxiv.2211.15089} & prompt completion and infilling, machine translation & classifier-free conditional & CDCD & MassiveText, C4, WMT2014, WMT2020 \\ \hline

\rowcolor{LightCyan}
\newcite{https://doi.org/10.48550/arxiv.2212.09412} &  machine translation and text summarization & classifier-free conditional & Difformer &  IWSLT14, WMT14, WMT16, Gigaword \\ \hline
	
\rowcolor{LightCyan}
\newcite{https://doi.org/10.48550/arxiv.2212.10325} & sequenceto-sequence text generation & classifier-free conditional & SeqDiffuSeq & CCD, Quasar-T, Wiki-Auto, QQP, IWSLT14 \\ \hline

\rowcolor{LightCyan}
\newcite{strudel2023selfconditioned} & conditional and unconditional text generation & unconditional, classifier-free guidance & SED & C4 \\ \hline

\end{longtable}

\clearpage
\onecolumn
\small\centering
\begin{longtable}{|p{1.8cm}|p{1.8cm}|p{1.6cm}|p{2cm}|p{2.8cm}|p{3cm}|}
\caption{This table summarizes the techniques each architecture uses or newly proposes for noise schedule and sampling, as well as the evaluation metrics that are applied. We also consider whether the architectures use the pre-trained models or not for further analysis. We use the following abbreviations for the sampling and evaluation metrics columns: PPL (Perplexity), MBR (Minimum Bayes Risk), bpc (bits per character), bpb (bits per raw byte), dist-1 (distinct unigram), div-4 (diverse 4-gram)}
\label{tab:table2} \\

\hline \multicolumn{1}{|c|}{\textbf{paper}} & \multicolumn{1}{c|}{\textbf{Architecture}} & \multicolumn{1}{c|}{\textbf{Pre-trained}} & \multicolumn{1}{c|}{\textbf{Schedule}} & \multicolumn{1}{c|}{\textbf{Sampling}} & \multicolumn{1}{c|}{\textbf{Evaluation Metrics}}  \\ \hline 
\endfirsthead

\multicolumn{ 6}{|c|}{\textbf{Discrete Diffusion Modes}} \\ \hline

\rowcolor{Gray} 
\newcite{austin2021structured} & D3PMs & No & $(T-t+1)^{-1}$ & & NLL, IS, FID, PPL \\ \hline

\rowcolor{Gray} 
\newcite{https://doi.org/10.48550/arxiv.2102.05379} & Multinomial Diffusion   & No & & & bpc, bpb  \\ \hline

\rowcolor{Gray} 
\newcite{https://doi.org/10.48550/arxiv.2112.06749} & SUNDAE & No & &  Low-temperature sampling, Argmax-unrolled decoding, Updating fewer tokens & BLEU  \\ \hline

\rowcolor{Gray} 
\newcite{https://doi.org/10.48550/arxiv.2211.15029} & DiffusionBERT & Yes & Spindle & x0-parameterization & PPL, BLEU, Self-BLEU \\ \hline

\rowcolor{Gray} 
\newcite{https://doi.org/10.48550/arxiv.2210.17432} & SSD-LM & No & Logits generation & Greedy projection, Sampling, Multi-hot & MAUVE, PPL, Dist-n \\ \hline

\rowcolor{Gray} 
\newcite{chen2023analog} & Bit Diffusion & No & & Self-Conditioning and Asymmetric Time Intervals & FID, BLEU-4, CIDEr, ROUGE-L \\ \hline

\rowcolor{Gray} 
\newcite{reid2023diffuser} & DiffusER & No & Edit-based Corruption & Beam Search, 2D Beam Search, Nucleus Sampling & BLEU, ROUGE  \\ \hline

\multicolumn{ 6}{|c|}{\textbf{Embedding Diffusion Modes}} \\ \hline

\rowcolor{LightCyan}
\newcite{https://doi.org/10.48550/arxiv.2205.14217} & Diffusion-LM & No & Cosine & Rounding Step and MBR & BLEU, word-level exact match, success rate, human evaluation \\ \hline

\rowcolor{LightCyan}
\newcite{https://doi.org/10.48550/arxiv.2210.08933} & DiffuSeq & No & Partial Noising & Classifier-free Conditional Denoising, MBR & BLEU, ROUGE, BERTScore, dist-1,  self-BLEU, div-4  \\ \hline

\rowcolor{LightCyan}
\newcite{https://doi.org/10.48550/arxiv.2211.15089} & CDCD & No & Time warping & Self-conditioning, Time warping & BLEU, AR-NLL, unigram entropy, MAUVE   \\ \hline

\rowcolor{LightCyan}
\newcite{https://doi.org/10.48550/arxiv.2212.09412} & Difformer & Yes &  Noise Factor & 2D parallel decoding & BLEU,  ROUGE-1/2/L, SacreBLEU \\ \hline
	
\rowcolor{LightCyan}
\newcite{https://doi.org/10.48550/arxiv.2212.10325} & SeqDiffuSeq & No & Adaptive noise schedule & Self-conditioning & BLEU, Rouge-L  \\ \hline

\rowcolor{LightCyan}
\newcite{strudel2023selfconditioned} & SED & Yes & Cosine & Self-conditioning & AR-NLL, unigram entropy \\ \hline

\end{longtable}

\section*{References}
\bibliography{anthology,custom}
\bibliographystyle{acl_natbib}